\documentclass[sigconf]{acmart}

\AtBeginDocument{%
  }

\usepackage{epsfig}
\usepackage{graphicx}
\usepackage{amsmath}
\usepackage{newtxtext,newtxmath}
\usepackage{booktabs}
\usepackage{multirow}
\usepackage{pifont}
\usepackage{url}
\usepackage{balance}

\newcommand{\etal}{{\it et al.}}
\newcommand{\eg}{{\it e.g.}}
\newcommand{\ie}{{\it i.e.}}

\copyrightyear{2023} 
\acmYear{2023} 
\setcopyright{acmlicensed}\acmConference[CIKM '23]{Proceedings of the 32nd ACM International Conference on Information and Knowledge Management}{October 21--25, 2023}{Birmingham, United Kingdom}
\acmBooktitle{Proceedings of the 32nd ACM International Conference on Information and Knowledge Management (CIKM '23), October 21--25, 2023, Birmingham, United Kingdom}
\acmPrice{15.00}
\acmDOI{10.1145/3583780.3614838}
\acmISBN{979-8-4007-0124-5/23/10}

\begin{document}

\title{Deep Variational Bayesian Modeling of Haze Degradation Process}

\author{Eun Woo Im}
\authornotemark[1]
\affiliation{%
  \institution{Department of Artificial Intelligence\\
  Hanyang University}
  \city{Seoul} \country{Republic of Korea}
}
\email{iameuandyou@hanyang.ac.kr}

\author{Junsung Shin}
\affiliation{%
  \institution{Department of Artificial Intelligence\\Hanyang University}
  \city{Seoul} \country{Republic of Korea}
}
\email{junsung6140@hanyang.ac.kr}
\authornote{Equal contribution.}

\author{Sungyong Baik}
\affiliation{%
  \institution{Department of Data Science\\
  Hanyang University}
  \city{Seoul} \country{Republic of Korea}
}
\email{dsybaik@hanyang.ac.kr}

\author{Tae Hyun Kim}
\affiliation{%
  \institution{Department of Computer Science\\
  Hanyang University}
  \city{Seoul} \country{Republic of Korea}
}
\email{taehyunkim@hanyang.ac.kr}
\authornote{Corresponding author.}


\begin{abstract}
Relying on the representation power of neural networks, most recent works have often neglected several factors involved in haze degradation, such as transmission (the amount of light reaching an observer from a scene over distance) and atmospheric light.
These factors are generally unknown, making dehazing problems ill-posed and creating inherent uncertainties.
To account for such uncertainties and factors involved in haze degradation, we introduce a variational Bayesian framework for single image dehazing.
We propose to take not only a clean image and but also transmission map as latent variables, the posterior distributions of which are parameterized by corresponding neural networks: dehazing and transmission networks, respectively.
Based on a physical model for haze degradation, our variational Bayesian framework leads to a new objective function that encourages the cooperation between them, facilitating the joint training of and thereby boosting the performance of each other.
In our framework, a dehazing network can estimate a clean image independently of a transmission map estimation during inference, introducing no overhead.
Furthermore, our model-agnostic framework can be seamlessly incorporated with other existing dehazing networks, greatly enhancing the performance consistently across datasets and models.
\end{abstract}

\ccsdesc[500]{Computing methodologies~Computational photography}

\keywords{Image dehazing; Variational Bayesian method; Computer vision; Machine learning; Image processing}


\maketitle

\begin{figure}[t]
    \centering
    \includegraphics[width=0.5\textwidth]{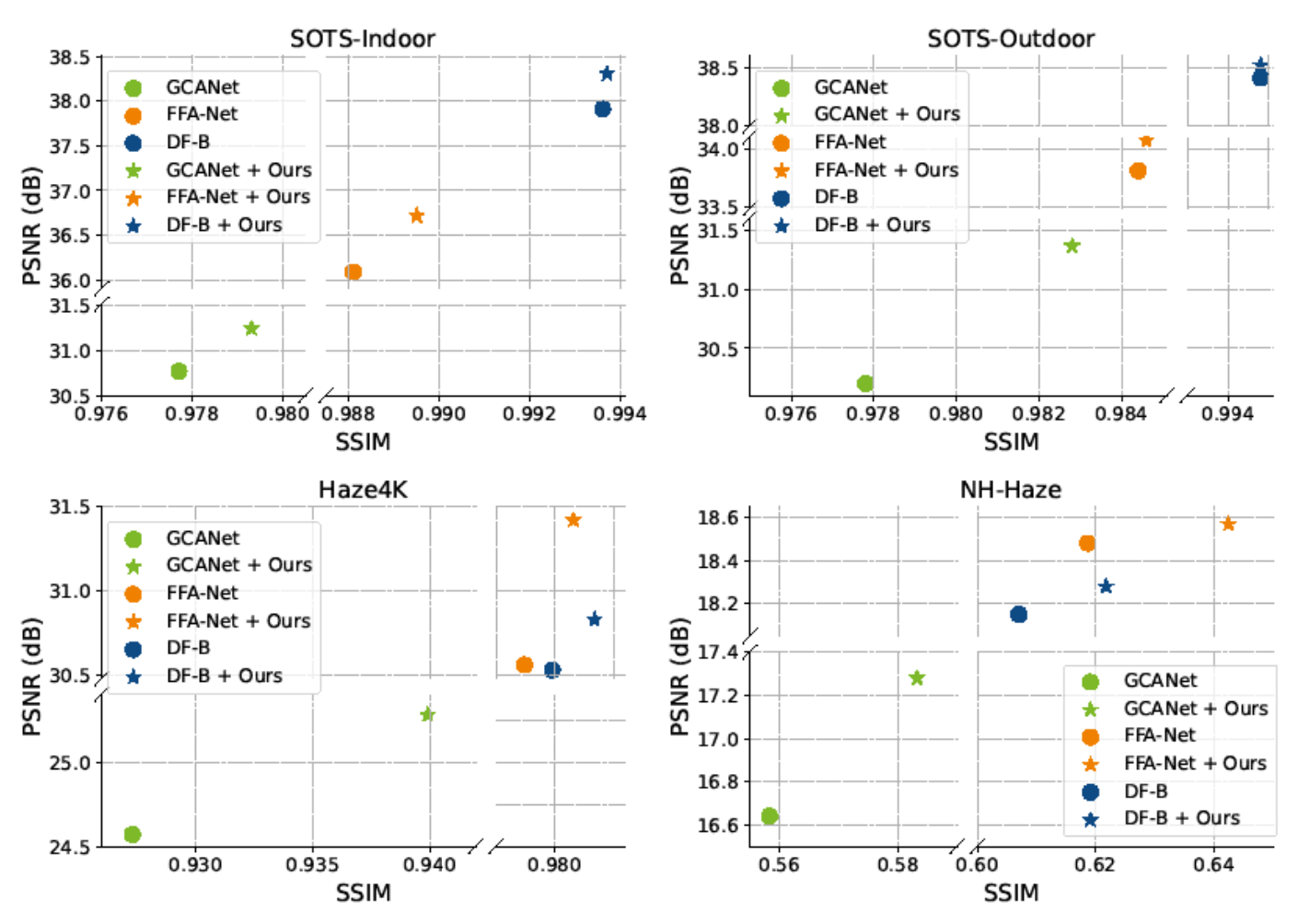}
    \caption{Our variatonal Bayesian framework is model-agnostic, and consistently improves the performance of existing dehazing neural networks across different benchmark datasets (SOTS~\cite{li2018benchmarking}, Haze4K~\cite{haze4k} and NH-Haze~\cite{Ancuti_2020_CVPR_Workshops})
    in terms of PSNR and SSIM values.
    Upward-right movement of the star indicates better restoration.
    }
    \label{fig:intro}
    \vspace{-2em}
\end{figure}

\vspace{-0.5em}
\section{Introduction}
Haze is an atmospheric phenomenon, where airborne particles (\eg, fog, dust, etc.) between the scene and an observer obscure the scene.
Such phenomenon causes poor visibility and thereby severely affects the performance of high-level vision tasks, such as semantic segmentation and object detection.
The extent of haze effects is determined by how far the scene is and the amount of airborne particles that either attenuate the visibility of the scene or scatter global atmospheric light towards an observer.
As such, an atmospheric scattering model~\cite{narasimhan2000chromatic, narasimhan2002vision} formulates haze effects as:
\begin{equation}
    I=J\odot t+A\cdot(1-t),
    \label{physical}
\end{equation}
where $I$ and $J$ are an observed hazy image and a scene radiance (\ie clean haze-free image), and $\odot$ indicates the pixel-wise multiplication.
The scalar $A$ denotes global atmospheric light, and $t$ is the transmission map representing the remaining fraction of light that reaches an observer from the scene.
In general, $t$ and $A$ are unknown, and thus recovering the clean image $J$ from a given hazy image $I$ is a highly ill-posed and challenging problem.

Based on this physical haze model, early works have imposed constraints with strong assumptions or priors (\eg, hazy regions have higher intensity values than haze-free regions~\cite{tan2008visibility} or haze-free regions have at least one color channel with low intensity~\cite{dcp}).
Due to such strong priors, these prior-based methods fail to work under scenarios where assumptions do not hold, resulting in poor generalization.
To alleviate this, recent data-driven approaches rely on large-scale datasets and the representation power of neural networks to recover clean haze-free images by learning to estimate transmissions~\cite{dehazenet, ren2016single} or directly learning a mapping of haze-free images~\cite{ren2018gated, zhang2018multi, mei2018progressive, liu2019dual, dudhane2019ri} or jointly estimate both from hazy images~\cite{zhang2018densely, zhang2019joint}.
However, there are inherent ambiguities and uncertainties (\eg, airlight-albedo ambiguity: we cannot tell how much light is from scene radiance or atmospheric light~\cite{fattal2008single}), causing inaccurate estimation of transmission map or haze-free images.

In this work, instead of focusing on building elaborate and effective network architectures, we shift the attention to modeling uncertainties involved in the single image dehazing problem.
To this end, we propose a new variational Bayesian framework, which incorporates the physical model (\ie, atmospheric scattering model) and takes not only hazy image and scene radiance but also transmission as latent variables conditioned on a given hazy image.
Then, the variational posteriors of scene radiance and transmission are parameterized by dehazing network and transmission network, respectively.
Upon our variational framework, we derive a new objective function that induces synergy between the training of two networks, thereby improving the overall dehazing performance.
Note that our framework allows for the joint training of dehazing and transmission networks, without making them dependent on each other.
Thus, during inference, a dehazing network can be used independently of a transmission network, introducing no extra overhead.
Furthermore, our framework design is model-agnostic, allowing for seamless integration with any dehazing neural network.
The contributions of this work can be summarized as follows:
\begin{enumerate}
    \item
    The proposed method models uncertainties of transmission maps and haze-free images by integrating the Bayesian modeling and data-driven methods.
    \item
    Our model-agnostic framework can seamlessly employ any conventional dehazing neural network without any architecture modification.
    \item
    Our framework consistently improves the performance of existing methods, including state-of-the-art models, across various benchmark datasets as illustrated in Figure~\ref{fig:intro}.
    
\end{enumerate}

\section{Related Work}
\label{sec:relatedwork}

In general, the haze effect is dependent on depth (\ie, a deep scene or a distant object produce minimal transmission, therefore resulting in a substantially hazy image).
The main challenge of the dehazing task lies in effectively extracting information on atmospheric light and transmission map merely from the given hazy image.
Most of existing dehazing strategies can be categorized into two approaches: prior and learning-based methods.

\paragraph{\textbf{Prior-based Methods}}

Early dehazing algorithms mostly depend on Eq.~(\ref{physical}) and statistical prior to impose constraints on the solution space.
Fattal~\etal~\cite{fattal2008single} assumed that shading of the object and transmission are statistically uncorrelated over the entire image patch to estimate the transmission map and albedo of the medium.
Tan~\etal~\cite{tan2008visibility} proposed to compute transmission map using Markov random field, utilizing the prior that haze-free regions have higher contrast than hazy ones.
He~\etal~\cite{dcp} proposed dark channel prior to estimate transmission maps and atmospheric light, based on the observation that the lowest intensity among the color channels of natural outdoor images is close to zero due to factors, such as shadow or color patterns.
Zhu~\etal~\cite{colorattenuation} proposed a linear model that restores depth maps with assumed color attenuation prior which describes the relationship between the pixel intensity, saturation, and their differences.
Berman~\etal~\cite{berman2016non} introduced a non-local method with haze-line prior, which assumes that a few hundred distinct colors can successfully approximate the color of haze-free regions, forming compact clusters in RGB space.
These methods often fail to work due to strong priors and assumptions.

\paragraph{\textbf{Learning-based Methods}}

As deep learning technology and large scale open source datasets become increasingly procurable, data-driven learning-based methods have become prevalent~\cite{wu2021contrastive, qu2019enhanced, jiang2022boosting, yang2022self, zheng2021ultra, chen2021psd}.
In contrast to prior-based methods, these learning-based methods learn to map hazy images to haze-free images directly in an end-to-end manner.
Cai~\etal~\cite{dehazenet} proposed DehazeNet, an end to end modeling with CNN, and Ren \etal~\cite{ren2016single} introduced, a multi-scale neural network (MSCNN), both successfully estimate transmission maps.
Li~\etal~\cite{aod} introduced a new variable dependent on hazy input by using the atmospheric scattering model in Eq.~(\ref{physical}), enabled to reconstruct latent clean image by predicting the variable.
Zhang~\etal~\cite{zhang2018densely} proposed an edge-preserving loss, multi-level architectures, and introduced a discriminator to estimate transmission map and atmospheric light mutually. 
Ren~\etal~\cite{ren2018gated} proposed several pre-processing steps and multi-scale-fusion-based network to learn confidence maps for improved global visibility. 
Dong~\etal~\cite{msbdn} incorporated generic recursive boosting algorithm in the dense feature fusion model for information preservation and performance improvement. 
Guo~\etal~\cite{guo2022image} added geometrical information to a transformer module and concatenated with a CNN module to increase the local and global connectivity.

\begin{figure*}
    \centering
    \includegraphics[width=1\textwidth]{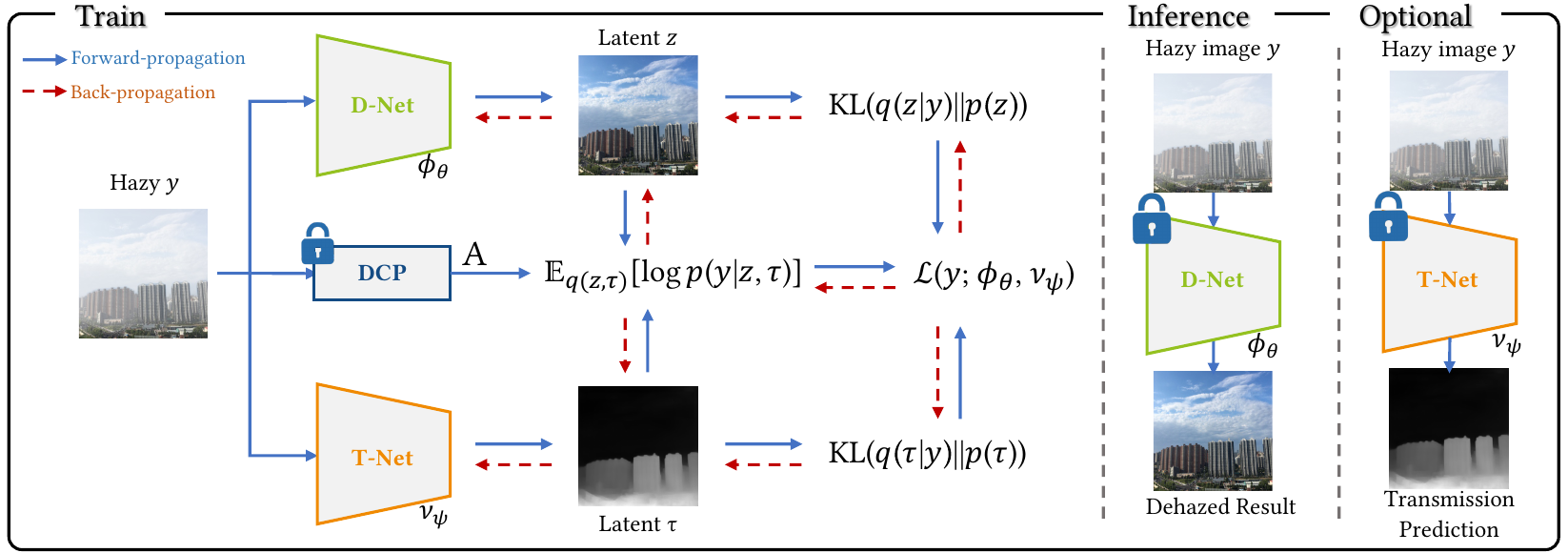}
    \caption{The architecture of proposed variational network. Blue solid lines represent forward process and red dotted lines denote gradient flow in back-propagation.
    Note that our \emph{D-Net} and \emph{T-Net} are not depending on specific network architectures.
    In addition, we can only employ \emph{D-Net} to output the haze-free image in the inference stage, hence no additional overhead during inference.}
    \label{fig:structure}
    \vspace{-1em}
\end{figure*}

\paragraph{\textbf{Variational Bayesian Modeling}}
Bayesian modeling allows for the modeling of uncertainties and latent variables that may not be readily apparent from the observed data.
While variational inference is a potent tool for approximating intricate probability distributions, its practical application requires careful consideration of prior knowledge.
For instance, to tackle denoising problem, Yue~\etal~\cite{yue2019variational} model noise-free image and its variance as latent variables utilizing a conjugate prior.
Wang~\etal~\cite{wang2021deep} leverage Dirichlet distribution to model blur kernel and deblurred image as latent variables and introduced two inference structures that are independent and dependent on the estimated blur kernel under blur process.
In this work, we take prior knowledge from the atmospheric scattering model, in which haze degradation involves several factors, such as transmission, that introduce uncertainties.
Motivated by the physical model, we employ variational Bayesian modeling and take not only hazy image and haze-free images but also transmission as latent for facilitating training and the modeling of uncertainties.

\vspace{-0.5em}
\section{Variational Haze Removal Framework}
\label{sec:method}

Let $\mathbb{D}$ be a training dataset composed of $n$ triplets $(x,y,t)$ of hazy image $y \in \mathbb{R}^{h \times w \times 3}$, ground truth clean image $x \in \mathbb{R}^{h \times w \times 3}$, and transmission map $t \in \mathbb{R}^{h \times w \times 1}$, respectively. 
$h$ and $w$ are the height and width of an image in RGB space, respectively.
Moreover, we denote latent haze-free image, latent transmission, and atmospheric light as $z$, $\tau$, and $A$.
We consider the clean image and transmission map as latent variables, and proceed to calculate their posterior distribution based on haze degradation.
This work aims to construct a variational function approximation of the posterior given a single hazy image through the Bayesian model including likelihood and priors.
Learning the joint distribution of our latent variables can further escalate the performance of the conventional dehazing networks.
The details are elaborated as in the following.

\subsection{Bayesian Model Construction}
\label{sec:Bayesian}

\paragraph{\textbf{Likelihood Model}}
Based on the atmospheric scattering model in Eq.~(\ref{physical}),
we start with taking intensity value of a hazy image as latent, which we assume to follow a Gaussian distribution as:

\begin{equation}
    y_i \sim \mathcal{N}(z_i \odot \tau_i+A(1-\tau_i), \sigma^2),
    \label{eq:hazy}
\end{equation}
where $y_i$, $z_i$, and $\tau_i$ denote pixel values of a hazy image, haze-free image, and transmission map at a pixel location $i$ respectively. 
Moreover, $\mathcal{N}(\mu, \sigma^2)$ is the Gaussian distribution with mean $\mu$ and variance $\sigma^2$.
For the sake of analytical feasibility and the basic properties of the Gaussian distribution that facilitate the parameterization of latent variables, we model $y_i$ as a Gaussian distribution~\cite{murphy2012machine}.
Since our training dataset $\mathbb{D}$ includes the ground truth haze-free image and transmission map, we can further take $z$ and $\tau$ as latent and train neural networks to estimate their posteriors.

\paragraph{\textbf{Haze-free Image}}

In general, optimizing the $L_1$ loss function encourages the median estimation of the observations rather than mean as with $L_2$ loss~\cite{huber1964robust, stigler1986history}.
Therefore, conventional dehazing networks favor $L_1$ loss variants to $L_2$ loss~\cite{qin2020ffa, guo2022image, song2022vision, lim2017enhanced} and minimize the absolute difference between the ground truth clean and predicted dehazed images during training to produce sharper edges/boundaries while suppressing the noise in homogeneous regions.
If regression model errors are assumed to follow a Laplace distribution, then maximum likelihood estimates of the distribution parameters correspond to $L_1$ regression estimates~\cite{meyer2021alternative}.
Accordingly, we model the haze-free image under data-driven Laplace prior as:
\begin{equation}
    z_i \sim {\rm Laplace}(x_i, \varepsilon_1^2),
    \label{eq:clean}
\end{equation}
where $x_i$ is a pixel value of the ground truth clean image at $i$, ${\rm Laplace}(n, \delta^2)$ denotes the Laplace distribution with parameters of location $n$ and scale $\delta^2$.
$\varepsilon_1^2$ is the mean absolute deviation from the median of $z_i$.
Given that $x_i$ serves as a reliable prior for $z_i$, we set a small value to $\varepsilon_1^2$.

\paragraph{\textbf{Transmission Map.}}

In this work, we assume that the atmosphere is homogeneous in the scene as in previous arts~\cite{cozman1997depth, narasimhan2002vision}.
Under this assumption, the scene radiance is exponentially attenuated~\cite{tan2008visibility} and the transmission map can be formulated with scattering coefficient $\beta$ and depth map $d$~\cite{cozman1997depth, nayar1999vision} as:
\begin{equation}
    t = e^{-\beta \cdot d}.
\end{equation}

For reason similar to Eq.~\eqref{eq:hazy}, we can model the probability of scene depth as a normal distribution.
When the logarithm of a variable is normally distributed, then the variable has log-normal distribution.
In addition, as in haze-free image modeling, a large number of transmission maps $t$ from the training data provide a strong data-driven prior to our latent transmission map $\tau$.
Therefore, we model the latent transmission map at pixel location $i$ as follows:
\begin{equation}
    \tau_i \sim {\rm Lognormal}(-\beta d_i, \varepsilon_2^2),
    \label{eq:trans}
\end{equation}
where $d_i$ denotes the pixel value of the latent depth map,
${\rm Lognormal}(m, b)$ is lognormal distribution with parameters of scale $m$ and scatter $b$.
Notably, $-\beta d_i$ and $\varepsilon_2^2$ are equal to $\ln t_i$ and $\varepsilon_d^2 \beta^2$, which shall be also a small value, similar to that of $\varepsilon_1^2$. We can compute $-\beta d_i$ as:
\begin{equation}
    -\beta d_i = \log \left( \frac{y_i - A}{x_i - A} \right ).
    \label{eq:logt}
\end{equation}

Moreover, to simplify the overall framework, we deal with $\sigma^2$, $\varepsilon_1^2$, $\varepsilon_2^2$ as hyper-parameters rather than latent variables, each of which controls the uncertainty of its associated variable.

\paragraph{\textbf{Atmospheric Light.}}
\label{para:atmospheric}

As for atmospheric light $A$, 
we adopt atmospheric light estimation result from the dark channel prior (DCP)~\cite{dcp} under the assumption of homogeneous atmosphere. 
As several conventional datasets provide the ground truth $A$, we can also treat $A$ as a latent and train a model to estimate it in our Bayesian framework. 
However, we empirically observed that the difference between dehazing results with the ground truth $A$ and the atmospheric light obtained by DCP are insignificant.
Therefore, we do not explicitly model the atmospheric light as latent and reduce the complexity of our Bayesian modeling by using the atmospheric light results by DCP in our work.

To be specific, the dark channel is defined as the morphological minimum filtered values among the RGB channels.
The most haze-opaque region in the image can be detected by collecting the top $0.1\%$ of brightest pixels in the dark channel~\cite{tan2008visibility}.
The one with the highest intensity in the corresponding hazy $y$ is selected as $A$.

\subsection{Variational Formulation of Posterior}

We aim to infer the posterior of the latent variables by merging the Bayesian models in Eqs.~(\ref{eq:hazy}-\ref{eq:trans}).
The direct estimation of the true posterior of the latent variables $z$ and $\tau$ solely from a single hazy image $y$ (\ie, $p(z,\tau|y)$) is computationally infeasible.
Therefore, we construct a variational surrogate distribution $q(z,\tau|y)$ to approximate $p(z,\tau|y)$.
Following the mean field assumption, we partition the variables into independent parts (\ie, assume the conditional independence between two variables $z$, and $\tau$):
\begin{equation}
    q(z,\tau|y) = q(z|y)q(\tau|y).
    \label{mean-field assumption}
\end{equation}
Using Eqs.~\eqref{eq:clean} and \eqref{eq:trans} with an assumption that surrogate distribution $q(z|y)$ and $q(\tau|y)$ have scale parameter $\varepsilon_1^2$ and scatter parameter $\varepsilon_2^2$ respectively, we formulate the variational posterior as:
\begin{equation}
    \begin{split}
        q(z|y) = &\prod_i \mathcal{\rm Laplace}(\phi_\theta(y)_i, \varepsilon_1^2),\\
        q(\tau|y) = &\prod_i \mathcal{\rm Lognormal}(\log \nu_\psi(y)_i, \varepsilon_2^2),
    \end{split}
    \label{variational posterior}
\end{equation}
where $\phi_\theta(\cdot)$, and $\nu_\psi(\cdot)$ are neural networks that are trained to estimate the posterior distribution parameters of latent variable $z$, and $\tau$, conditioned on the input hazy image $y$.
Since our proposed framework focuses on modeling the posterior of latent variables, without any assumption on the form of $\phi_\theta(\cdot)$, and $\nu_\psi(\cdot)$, our framework is model-agnostic.
In particular, $\phi_\theta$, which we call \emph{D-Net}, can be any existing dehazing networks (\eg,~\cite{chen2018gated}), which is a neural network with parameters $\theta$ trained to estimate the haze-free image.
Similarly, $\nu_\psi$, named as \emph{T-Net}, is an auxiliary neural network with parameters $\psi$ to estimate the transmission map from a given input hazy image.
Note that there is no additional overhead during inference as we can estimate a haze-free image with only \emph{D-Net}, and the transmission map is optionally obtainable as shown in Figure~\ref{fig:structure}.

\subsection{Variational Lower Bound}
\label{sec:vlb}
With the functional parameterization of the variational posterior, we can optimize the trainable parameters $\theta$, and $\psi$ to maximize the posterior probability.
To do so, we decompose the marginal log-likelihood and obtain the variational lower bound.
For notational simplicity, we use $\phi_i$ and $\nu_i$ rather than  $\phi_\theta(y)_i$ and $\nu_\psi(y)_i$, then the marginal log-likelihood is given by,
\begin{equation}
    \begin{split}
    & \log p(y;z,\tau) = \mathbb{E}_{q(z,\tau|y)}\left[ \log p(y) \right] \\
    & \quad = \iint q(z,\tau|y) \log \left( \frac{p(y, z, \tau)}{p(z, \tau | y)} \right) \, dz d\tau \\
    & \quad = \iint q(z,\tau|y) \log \left( \frac{p(y|z,\tau) p(z) p(\tau)}{p(z,\tau|y)} \right) \, dz d\tau \\
    & \quad = \iint q(z,\tau|y) \log \left(\frac{p(y| z,\tau) p(z) p(\tau)}{q(z,\tau|y)} \frac{q(z,\tau|y)}{p(z,\tau|y)} \right) \, dz d\tau \\
    & \quad = \iint q(z,\tau|y) \log \left(\frac{p(y|z,\tau) p(z) p(\tau)}{q(z,\tau|y)} \right) \, dz d\tau  \\
    & \qquad + \iint q(z,\tau|y) \log \left(\frac{q(z,\tau|y)}{p(z,\tau|y)} \right) \, dz d\tau \\
    & \quad = \mathbb{E}_{q(z,\tau|y)} \left[\log\left(\frac{p(y|z,\tau) p(z) p(\tau)}{q(z,\tau|y)} \right) \right]  + {\rm KL} \left( q(z, \tau | y) \| p(z, \tau | y) \right) \\
    & \quad \equiv \mathcal{L}(y;\phi_\theta, \nu_\psi) + {\rm KL}\left( q(z, \tau | y) \| p(z, \tau | y) \right).
    \end{split}
    \label{likelihood decomposition}
\end{equation}    
where ${\rm KL}(\cdot \| \cdot)$ computes the Kullback–Leibler (KL) divergence of two distributions, and $\mathcal{L}(y;\phi_\theta, \nu_\psi)$ is the variational lower bound which can be combined with Eqs.~(\ref{mean-field assumption}) and (\ref{variational posterior}) as follows:
\begin{equation}
    \begin{split}
    & \mathcal{L}(y;\phi_\theta, \nu_\psi) = \mathbb{E}_{q(z,\tau|y)} \left[ \log \left( \frac{p(y|z,\tau)p(z)p(\tau)}{q(z,\tau|y)} \right) \right] \\
    & \quad = \mathbb{E}_{q(z,\tau|y)} \left[ \log p(y|z,\tau) \right] - \mathbb{E}_{q(z,\tau|y)} \left[ \log \left( \frac{p(z)p(\tau)}{q(z|y)q(\tau|y)} \right) \right] \\
    & \quad = \mathbb{E}_{q(z,\tau|y)} \left[ \log p(y|z,\tau) \right] - {\rm KL} \left( q(z|y)\| p(z) \right) - {\rm KL} \left( q(\tau|y) \| p(\tau) \right).
    \end{split}
    \label{eq:loss}
\end{equation}
and each term in Eq.~\eqref{eq:loss} can be calculated analytically as follows: 
\begin{equation}
    \begin{split}
    & \mathbb{E}_{q(z,\tau|y)}\left[\log p(y|z,\tau)\right] \\
    & = \sum_{i=1}^{hw}\left\{-\frac{\log 2 \pi \sigma^2}{2} - \frac{(y_i - (\phi_i\nu_i+A(1- \nu_i)))^2 + \sigma^2}{2\sigma^2}\right\},
    \end{split}
    \label{eq:likelihood}
\end{equation}
\begin{equation}
    \begin{split}
    & {\rm KL}(q(z|y)\|p(z)) \\
    & = \sum_{i=1}^{hw} \left \{ \exp \left(- \frac{|\phi_i - x_i|}{\varepsilon_1^2} \right) + \frac{{ |\phi_i - x_i|}}{\varepsilon_1^2} - 1 \right\},
    \end{split}
    \label{eq:kl_laplace}
\end{equation}
and
\begin{equation}
    {\rm KL}(q(\tau|y)\|p(\tau)) = \sum_{i=1}^{hw} \left\{ \frac{1}{2\varepsilon_2^2} (\log \nu_i - \log t_i)^2  \right \}.
    \label{eq:kl_lognormal}
\end{equation}
Note that all terms in Eq.~\eqref{eq:loss} are differentiable, and we can train the network parameters $\theta$ and $\psi$ over the given training dataset $\mathbb{D}$ by optimizing the following objective function:
\begin{equation}
    \min_{\theta, \psi} -  \mathcal{L}(y ; \phi_\theta, \nu_\psi).
    \label{eq:total_loss}
\end{equation}

\begin{table*}[t]
  \centering
  \begin{tabular}{l c c c c c c}
    \toprule
    \multirow{2}{*}{Methods} & \multicolumn{2}{c}{SOTS-Indoor} & \multicolumn{2}{c}{SOTS-Outdoor} & \multicolumn{2}{c}{Haze4K} \\
    \cline{2-7}
    & PSNR$\uparrow$ & SSIM$\uparrow$ & PSNR$\uparrow$ & SSIM$\uparrow$ & PSNR$\uparrow$ & SSIM$\uparrow$ \\
    \midrule
    DCP~\cite{dcp}                  & 16.62 & 0.818 & 19.13 & 0.815 & 14.01 & 0.760 \\
    BCCR~\cite{meng2013efficient}   & 17.04 & 0.785 & 15.51 & 0.791 & - & -\\
    CAP~\cite{colorattenuation}     & 18.97 & 0.815 & 18.14 & 0.759 & 16.32 & 0.782 \\
    NLD~\cite{berman2016non}        & 17.29 & 0.777 & 17.97 & 0.819 & - & - \\
    DehazeNet~\cite{dehazenet}      & 19.82 & 0.821 & 24.75 & 0.927 & 19.12 & 0.840 \\
    AOD-Net~\cite{aod}              & 20.51 & 0.816 & 24.14 & 0.920 & 17.15 & 0.830 \\
    MSBDN~\cite{msbdn}              & 33.67 & 0.985 & 33.48 & 0.982 & 22.99 & 0.850 \\
    DeHamer~\cite{guo2022image}     & 36.63 & 0.988 & 35.18 & 0.986 & 27.28 & 0.956 \\
    \midrule
    GCANet~\cite{chen2018gated}     & 30.23 (30.77) & 0.9764 (0.9777) & - (30.20) & - (0.9778) & - (24.57) & - (0.9273) \\
    GCANet + Ours                   & 31.25 & 0.9793 & 31.37 & 0.9828 & 25.28 & 0.9399 \\
    \hline
    FFA-Net~\cite{qin2020ffa}       & 36.39 (36.09) & 0.9886 (0.9881) & 33.57 (33.81) & 0.9839 (0.9844) & - (30.56) & - (0.9787) \\
    FFA-Net + Ours                  & 36.72 & 0.9895 & 34.07 & 0.9846 & \textbf{31.42} & 0.9808 \\
    \hline
    DehazeFormer-B~\cite{song2022vision} & 37.84 (37.91) & 0.9936 (0.9936) & 34.95 (38.41) & 0.9843 (\textbf{0.9948}) & - (30.53) & - (0.9799) \\
    DehazeFormer-B + Ours & \textbf{38.31} & \textbf{0.9937} & \textbf{38.52} & \textbf{0.9948} & 30.83 & \textbf{0.9817} \\
    \bottomrule
  \end{tabular}
  \caption{
  The PSNR(dB), SSIM comparison of image dehazing methods on different synthetic data benchmarks.
  The numbers within parenthesis represent reproduced results.
  Baseline dehazing networks are GCANet~\cite{chen2018gated}, FFA-Net~\cite{qin2020ffa}, and basic DehazeFormer~\cite{song2022vision}.
  + Ours indicates baseline networks trained with the proposed Bayesian framework.
  The best values are indicated as bold text.
  }
  \vspace{-2.5em}
  \label{tab:benchmark}
\end{table*}

\subsection{Learning with Variational Lower Bound}
\label{sec:discussions}
By minimizing the final objective function in Eq.~\eqref{eq:total_loss} through conventional back-propagation without using the reparameterization trick~\cite{kingma2013auto}, we can train the parameters of networks $\phi_\theta$ and $\nu_\psi$ and estimate the posterior of latent variables $z$ and $\tau$ as illustrated in Figure~\ref{fig:structure}.
Notably, the roles of three terms composing the total objective can be explained as follows.
The first term represents the likelihood of the observed hazy images and is responsible for encouraging cooperation between dehazing and transmission networks based on the Eq.~\eqref{physical}, which describes the relationship between the three latent variables: hazy image, haze-free image, and transmission.
The second (Eq.~\eqref{eq:kl_laplace}) and the third (Eq.~\eqref{eq:kl_lognormal}) terms act as regularization, making the posterior distribution close to prior distribution.
Therefore, two separate networks $\phi_\theta$ and $\nu_\psi$ can complement each other with the aid of the joint term, and they are simultaneously trained by simulating the physical haze degradation process.

Furthermore, $\sigma^2$, $\varepsilon_1^2$ and $\varepsilon_2^2$ can be interpreted as not only uncertainty of each variable but also the importance of the associated term.
For instance, the importance of the KL divergence between $q(z|y)$ and $p(z)$ increases as $\varepsilon_1^2$ approaches to zero.

\section{Experimental Results}
\label{sec:experiment}

\subsection{Experimental Setting}
\label{sec:exp_setting}
\paragraph{\textbf{Datasets}}
We conducted our experiments on both synthetic and real-world datasets.
We utilize the RESIDE~\cite{li2018benchmarking} and Haze4K~\cite{haze4k} as synthetic datasets, and the NH-Haze~\cite{Ancuti_2020_CVPR_Workshops} and Fattal evaluation set~\cite{Fattal2014} for real-world datasets.
The RESIDE benchmark comprises synthetic hazy images along with their corresponding clean images captured in both indoor and outdoor scenarios.
The Synthetic Objective Test Set (SOTS) is used to evaluate the performance of the models on RESIDE dataset.
The indoor training set (ITS) of RESIDE benchmark consists of 13990 generated hazy images from 1399 clean images.
The outdoor training set (OTS) of RESIDE benchmark includes a total of 313950 hazy images generated by using the collected real outdoor images.
Haze4K is constructed by generating 4000 hazy images with randomly sampled atmospheric light $A$ and  scattering coefficient $\beta$ from 500 clean indoor images in NYU-Depth~\cite{silberman2012indoor} and 500 outdoor images in OTS.
NH-Haze contains 55 paired images of real-world haze scenes.

\paragraph{\textbf{Implementation}}
For our \emph{D-Net}, we can employ any conventional dehazing networks, and we use  GCANet~\cite{chen2018gated}, FFA-Net~\cite{qin2020ffa}, and a recent state-of-the-art network DehazeFormer-B~\cite{song2022vision} as our baseline dehazing networks. For our \emph{T-Net}, we use GCANet with a clamping activation function on the output layer.
For fair comparison, we follow all training and evaluation strategies of the baselines (\eg, total epoch, optimizer, etc.) and our Bayesian framework is implemented based on the officially available code of each baseline.

In the case of real-world NH-Haze dataset, total training epoch is set to 300 using the official train-test split.
As the NH-Haze train set lacks the ground truth transmission map, we estimated the map using a clean and hazy image pair while assuming $A=1$ from Eq.~(\ref{physical})
(\ie, $t=(I-A)/(J-A+\epsilon) \in \mathbb{R}^{h \times w \times 3}$ with $\epsilon=10^{-6}$ for numerical stability).

Notably, as our baseline networks originally employ either $L_1$ or $L_2$ loss functions, we do not employ additional objective functions (\eg, adversarial loss~\cite{zhao2020pyramid, dong2020fd, Li_2018_CVPR, du2019recursive}, contrastive loss~\cite{wu2021contrastive}, and perceptual loss~\cite{Liu_2019_ICCV, qu2019enhanced}) for fair comparisons.
We empirically determine $\sigma^2=10^{-5}$, $\varepsilon_1^2=10^{-6}$, and $\varepsilon_2^2=10^{-5}$.
Our source code is publically available.\footnote{\tt \href{https://github.com/imeunu/Variational-Dehazing-Networks}{https://github.com/eunwooim/Variational-Dehazing-Networks}}

\subsection{Performance Evaluation}
\label{sec:exp_result}
To evaluate the performance of the proposed Bayesian framework, we compare the dehazing results with and without using the proposed framework both on synthetic and real-world haze datasets.

\paragraph{\textbf{Results on Synthetic Datasets}}
Table~\ref{tab:benchmark} presents the dehazing results on three different datasets (SOTS-Indoor, SOTS-Outdoor, and Haze4K), comparing with DCP~\cite{dcp}, BCCR~\cite{meng2013efficient} CAP~\cite{colorattenuation}, NLD~\cite{berman2016non}, DehazeNet~\cite{dehazenet}, AOD-Net~\cite{aod}, MSBDN~\cite{msbdn}, and DeHamer~\cite{guo2022image}.
Notably, for our baseline GCANet~\cite{chen2018gated}, FFA-Net~\cite{qin2020ffa}, and DehazeFormer-B~\cite{song2022vision}, we provide two sets of PSNR and SSIM values: the scores reported in their original manuscripts and reproduced numbers in our experiments which are within the parentheses.
As shown, our proposed method integrated with DehazeFormer-B obtains the highest metric scores in every domain, except for the PSNR on Haze4K, where FFA-Net + Ours performs the best.

\begin{figure*}[t!]
    \centering
    \includegraphics[width=0.97\textwidth]{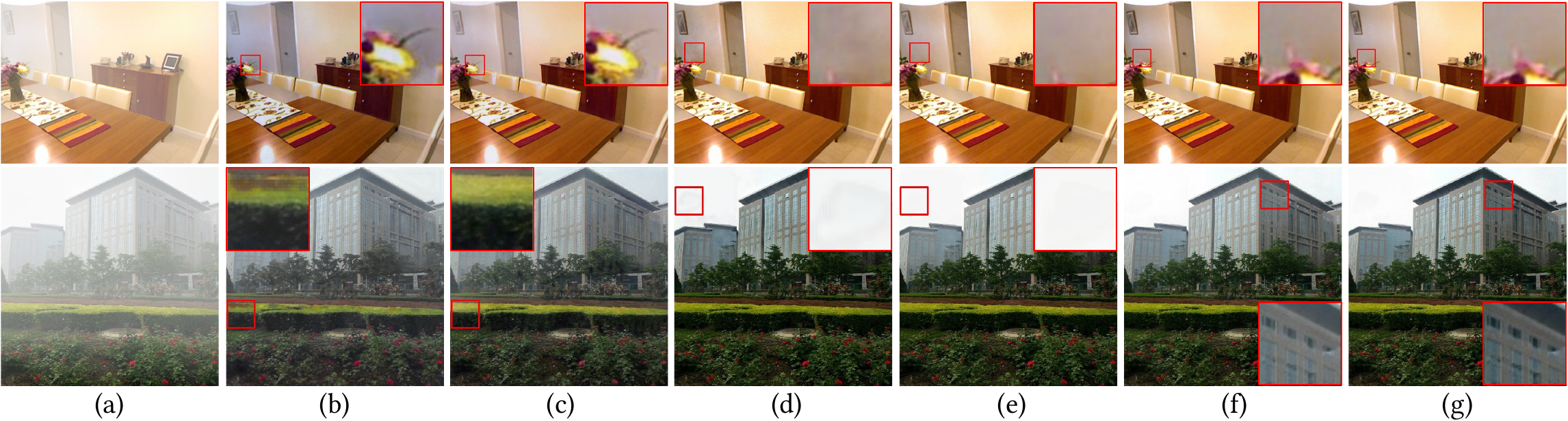}
    \caption{
    Visual comparisons between the baseline models and our enhanced models on the SOTS dataset~\cite{li2018benchmarking}.
    (a) Input hazy image.
    (b) GCANet~\cite{chen2018gated}.
    (c) GCANet + Ours.
    (d) FFA-Net~\cite{qin2020ffa}.
    (e) FFA-Net + Ours.
    (f) DehazeFormer-B~\cite{song2022vision}.
    (g) DehazeFormer-B + Ours.
    Best viewed on high-resolution display.
    }
    \label{fig:sota_vis}
    \vspace{-1em}
\end{figure*}

Moreover, in Figure~\ref{fig:sota_vis}, we present visual comparisons of our method with baseline models on SOTS-Indoor test set. 
We see that our method produces clear images with less artifacts.

\begin{table}
    \centering
    \begin{tabular}{l c c c c c }
      \toprule
        Method & PSNR$\uparrow$ & SSIM$\uparrow$ & LPIPS$\downarrow$ \\
        \hline
        DCP~\cite{dcp} & 12.72 & 0.4419 & 0.5168 \\ 
        BCCR~\cite{meng2013efficient} & 13.12 & 0.4831 & -\\
        CAP~\cite{colorattenuation} & 12.88 & 0.4929 & 0.6223 \\
        NLD~\cite{berman2016non}    & 12.23 & 0.4823 & - \\
        AOD-Net~\cite{aod} & 15.31 & 0.4584 & 0.5121 \\
        MSBDN~\cite{msbdn} & 17.34 & 0.5566 & 0.5026 \\
        DeHamer~\cite{guo2022image} & 17.91 & 0.5781 & 0.4816 \\
        \hline
        GCANet~\cite{chen2018gated} & 16.64 & 0.5583 & 0.4356 \\ 
        GCANet+Ours & 17.28 & 0.5832 & 0.4057\\
        \hline
        FFA-Net~\cite{qin2020ffa} & 18.48 & 0.6186 & 0.3694 \\
        FFA-Net + Ours & \textbf{18.57} & \textbf{0.6424} & \textbf{0.3415} \\
        \hline
        DehazeFormer-B~\cite{song2022vision} & 18.15 & 0.6070 & 0.4192 \\
        DehazeFormer-B + Ours & 18.28 & 0.6217 & 0.4066\\
        \bottomrule
        \end{tabular}
        \caption{
        The PSNR(dB), SSIM and LPIPS results on NH-Haze test set~\cite{Ancuti_2020_CVPR_Workshops}.
        }
    \vspace{-3em}
    \label{tab:nh_haze}
\end{table}

\paragraph{\textbf{Results on Real-world Datasets}}
In Table~\ref{tab:nh_haze}, the evaluation results in terms of PSNR, SSIM, and LPIPS obtained from the NH-Haze test set~\cite{Ancuti_2020_CVPR_Workshops} are provided.
We compare with DCP, CAP, MSBDN, and our baselines.
Our proposed method, when combined with FFA-Net, achieved the best performance in every metric.
Notably, we observed an improvement in performance of over 0.28 dB in PSNR and 0.021 in SSIM on average.

\begin{figure*}[t!]
    \centering
    \includegraphics[width=0.97\textwidth]{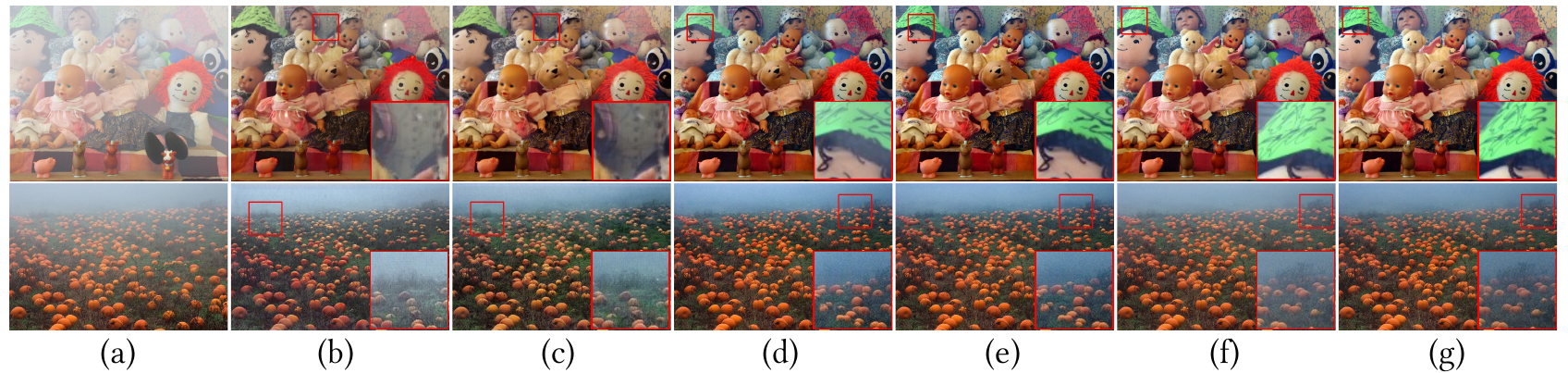}
    \caption{Visual comparisons of image dehazing methods on Fattal evaluation set \cite{Fattal2014}.
    (a) Hazy input image.
    (b) GCANet~\cite{chen2018gated}.
    (c) GCANet + Ours.
    (d) FFA-Net~\cite{qin2020ffa}.
    (e) FFA-Net + Ours.
    (f) DehazeFormer~\cite{song2022vision}.
    (g) DehazeFormer + Ours.
    }
    \label{fig:real}
    \vspace{-1em}
\end{figure*}

We compare our method against baseline methods on the Fattal evaluation set, as demonstrated in Figure~\ref{fig:real}, where all networks were trained on the Haze4K dataset.
It can be observed that models trained with our method outperform each baseline, and effectively removing haze while producing more vivid colors with less artifacts.
The results demonstrate the effectiveness of our framework in removing depth-independent haze, with manually calculated transmission map from atmospheric scattering model.

\paragraph{\textbf{User Study Results}}
Due to the lack of ground truth clean images or object annotations in the Fattal evaluation set~\cite{Fattal2014}, as well as the need for further evaluation of the qualitative aspects, we conducted a user study.
The details are described as follows.
First, we randomly selected six images each from SOTS-indoor, SOTS-outdoor~\cite{li2018benchmarking}, Haze4K test set~\cite{haze4k}, and Fattal evaluation set~\cite{Fattal2014}.
For each hazy image, we randomly chose a pair of dehazing results from one of the three base models (GCANet~\cite{chen2018gated}, FFA-Net~\cite{qin2020ffa}, and DehazeFormer-B~\cite{song2022vision}) and corresponding enhanced model by our approach.
To further validate the performance of perceptual quality in a real-world scenario, we selected images containing distinct objects on Fattal evaluation set and utilized Yolov5~\cite{glenn_jocher_2020_4154370} to detect the objects in the dehazed image pair.
Finally, 18 raters were asked to vote on 18 dehazed result pairs that appeared more visually convincing, and 6 object detection output pairs with more precise bounding boxes and accurate classification of the object.

\begin{figure}[h]
    \centering
    \includegraphics[width=0.5\textwidth]{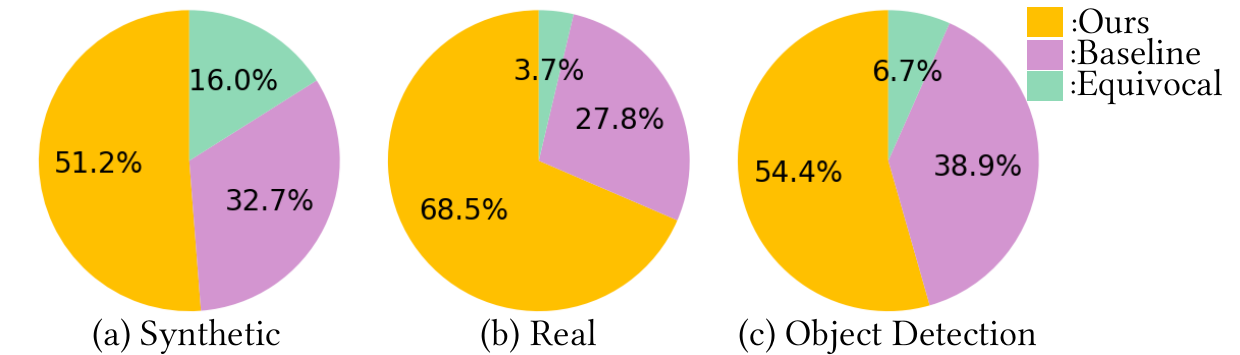}
    \caption{
    User study results.
    }
    \label{fig:user_study}
    \vspace{-1em}
\end{figure}

As summarized in Figure~\ref{fig:user_study}, the pie charts (a) and (b) indicate that our proposed method consistently generates more visually pleasing clean images than the baseline models.
In addition, pie chart (c) demonstrates that our framework produces dehazed images with superior perceptual quality.
Therefore, the results of our user study clearly demonstrate the effectiveness of our proposed method in improving the quality of dehazed images.

\subsection{Object Detection Application}
\label{sec:exp_qual}
We further assess the quality of the estimated haze-free images by evaluating how much the haze removal improves a downstream task: object detection in this work.
We perform experiments with YOLOv5~\cite{glenn_jocher_2020_4154370} as an object detector on KITTI Haze dataset~\cite{msbdn}, which is synthesized based on KITTI detection dataset~\cite{geiger2012we}, following the dataset generation algorithm of RESIDE dataset~\cite{li2018benchmarking} with depth estimation method~\cite{monodepth2}.
The quality of dehazed images is evaluated with how much detection accuracy improves in terms of mean average precision.

\begin{table}
    \centering
    \begin{tabular}{l c c}
    \toprule
    \textbf{YOLOv5}~\cite{glenn_jocher_2020_4154370} & mAP50$\uparrow$ & mAP50-95$\uparrow$ \\
    \midrule
    Hazy images & 68.5 & 45.2 \\
    Ground truth clean images & 93.0 & 69.6 \\
    \hline
    GCANet~\cite{chen2018gated} & 76.4 & 53.7 \\
    GCANet + Ours & 78.6 & 55.3 \\
    \hline
    FFA-Net~\cite{qin2020ffa} & 75.2 & 53.3 \\
    FFA-Net + Ours & 78.3 & 56.6 \\
    \hline
    DehazeFormer-B~\cite{song2022vision} & 87.3 & 63.2  \\
    DehazeFormer-B + Ours & \textbf{87.6} & \textbf{63.7} \\
    \bottomrule
  \end{tabular}
  \caption{Obejct detection results with YOLOv5~\cite{glenn_jocher_2020_4154370}.
  Mean average precision scores larger than 0.5 overlap IOU (mAP50) and 0.5$\sim$0.95 overlap IOU (mAP50-95) on the KITTI Haze dataset~\cite{msbdn} are reported.
  }
  \vspace{-3.25em}
  \label{tab:highlevel}
\end{table}

\begin{figure*}[t!]
    \centering
    \includegraphics[width=1\textwidth]{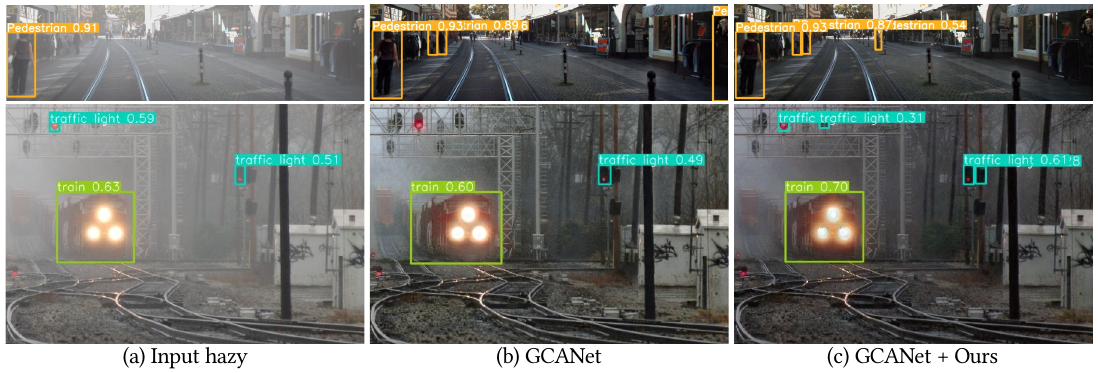}
    \caption{
    Object detection results by Yolov5~\cite{glenn_jocher_2020_4154370} on estimated clean images.
    \textbf{Top to bottom}: Results on the KITTI Haze dataset~\cite{msbdn} and the Fattal evaluation set~\cite{Fattal2014}.
    (a) Detection results from input hazy images.
    (b) Detection results with GCANet.
    (c) Detection results with GCANet + Ours.
    Best viewed on high-resolution display.}
    \label{fig:detection}
    \vspace{-1em}
\end{figure*}

Table~\ref{tab:highlevel} summarizes the detection performance on hazy images, ground truth clean images (upper bound), estimated clean images by the baselines GCANet~\cite{chen2018gated}, FFA-Net~\cite{qin2020ffa},  DehazeFormer-B~\cite{song2022vision}, and ours applied to each model, where all models are trained with RESIDE outdoor dataset~\cite{li2018benchmarking}.
The object detector clearly benefits from the haze removal, 
and is greatly improved by using the dehazed images, and we observe that our framework shows consistent improvement over baselines.
Moreover, Figure~\ref{fig:detection} presents the qualitative results of object detection both on synthetic dataset~\cite{msbdn} and on real-world dataset~\cite{Fattal2014} and demonstrates that our framework not only improves the dehazing performance but also allows detection module to recognize distant objects with more clear image, while enhancing the confidence.

\vspace{-0.5em}
\subsection{Ablation Study}
\label{sec:exp_abl}

\begin{figure}
    \centering
    \includegraphics[width=0.5\textwidth]{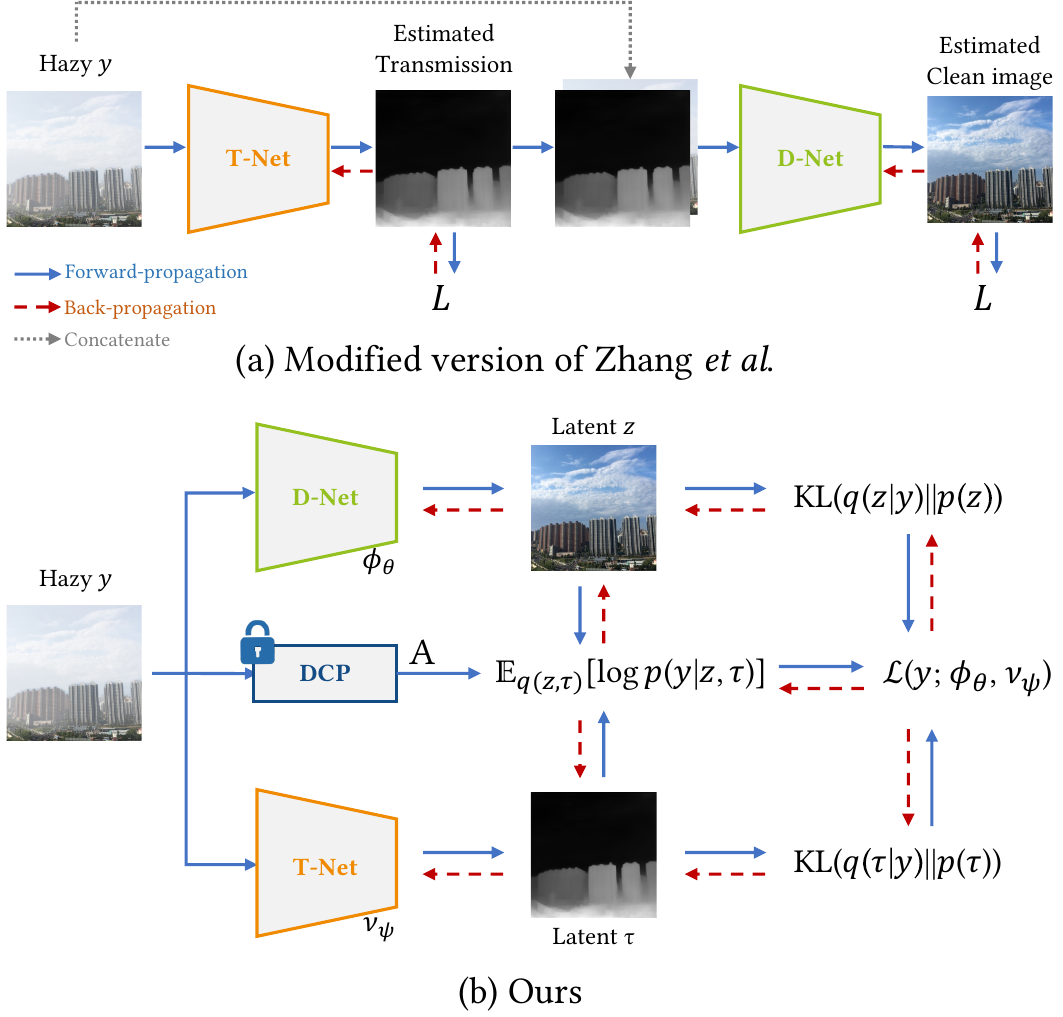}
    \captionof{figure}{
    Illustration of the modification made to the framework from Zhang~\etal~\cite{zhang2019joint}.
    }
    \label{fig:sup_modify_zhang}
    \vspace{-2.25em}
\end{figure}

\begin{table}
  \centering
  \begin{tabular}{c | c c | c c c}
    \toprule
     Configuration & Joint & Loss &  PSNR$\uparrow$ & SSIM$\uparrow$ \\
    \midrule
    GCANet~\cite{chen2018gated}    &\text{\ding{55}}   & $L_2$  & 24.57             & 0.9273 \\
    GCANet + Modified \cite{zhang2019joint} &\ding{51}          & $L_2$  & 24.97             & 0.9153 \\
    GCANet + Ours  &\ding{51}          & Eq.~\eqref{eq:total_loss}  & \textbf{25.28}    & \textbf{0.9399} \\
    \bottomrule
  \end{tabular}
  \caption{
  Comparison with different joint training method.
  The terms Joint and Loss indicate whether the training strategy is joint or not and the training objective, respectively.
  We compare original GCANet, GCANet with a modified joint training strategy from ~\cite{zhang2019joint} (GCANet + Modified~\cite{zhang2019joint}) and GCANet with our final Bayesian framework (GCANet + Ours).
  The PSNR and SSIM results are evaluated on Haze4K~\cite{haze4k} test set.
  }
  \vspace{-3em}
  \label{tab:ablation_joint}
\end{table}

\paragraph{\textbf{Joint optimization and Bayesian modeling}}
we performed ablation study in order to validate the effectiveness of our Bayesian modeling, we compared our framework with the slghtlty modified version of the joint training framework proposed by Zhang~\etal~\cite{zhang2019joint}.
Specifically, the estimated joint transmission map is concatenated to the input of the dehazing module, facilitating the joint training of the dehazing and transmission estimation module as illustrated in Figure~\ref{fig:sup_modify_zhang}.
Note that for a fair comparison, we have excluded the adversarial network, adversarial loss, and perceptual loss, and employed GCANet~\cite{chen2018gated} for both modules.

To verify the effectiveness of the joint optimization strategy and our Bayesian modeling, we compare these with the modified version of joint training framework introduced in Zhang~\etal~\cite{zhang2019joint}.
Specifically, the estimated transmission map is concatenated to the input of the dehazing module to jointly train the dehazing and transmission estimation networks.
We utilize GCANet for dehazing and transmission estimation module for a fair comparison and train with the ground truth clean image and transmission map using the $L_2$ objective.
This configuration (GCANet + Modified \cite{zhang2019joint}) allows joint training, but does not take into account the haze degradation process.
From the comparison result, it is observed that our final model (GCANet + Ours) using the joint optimization with our Bayesian framework outperforms the best.
We analyze that the likelihood term (\ie, Eq.~\eqref{eq:likelihood}) in the proposed objective is responsible for leveraging the relationships between latent variables and uncertainty, resulting in our method's superior performance in comparison with the modified joint method, which lacks this term.
In addition, joint optimization helps to utilize transmission information, thus both contribute to performance gain.

\paragraph{\textbf{Prior Distribution.}}
We further justify our choice of prior distributions for $z$ and $\tau$ by performing ablation study on the prior distributions.
As reported in Table~\ref{tab:distribution}, we perform ablations based on GCANet backbone, replacing each prior distribution with Gaussian distributions of variance ($\sigma^2$, $\varepsilon_1^2$, $\varepsilon_2^2$).
In the results, Laplace and Lognormal distribution tend to yield better performance.
Notably, using the Laplace distribution for $z$ provides higher SSIM values than Gaussian distribution, corroborating our motivation that Laplace encourages $z$ to learn the latent of sharp boundaries.

\subsection{Discussion}
\label{sec:tnet_explained}

\paragraph{\textbf{Role of \emph{T-Net}.}}
\emph{T-Net} serves as an auxiliary branch in our framework, designed to estimate the scale parameter of the Lognormal distribution modeling the latent variable $\tau$, which is not utilized in the inference phase.
However, we can further inspect the reason behind the output of \emph{D-Net} by examining the output of \emph{T-Net}.
For instance, the dehazing network may prioritize the restoration of more vivid colors in areas where lower transmission is estimated.

\begin{figure}[t!]
    \centering
    \includegraphics[width=0.5\textwidth]{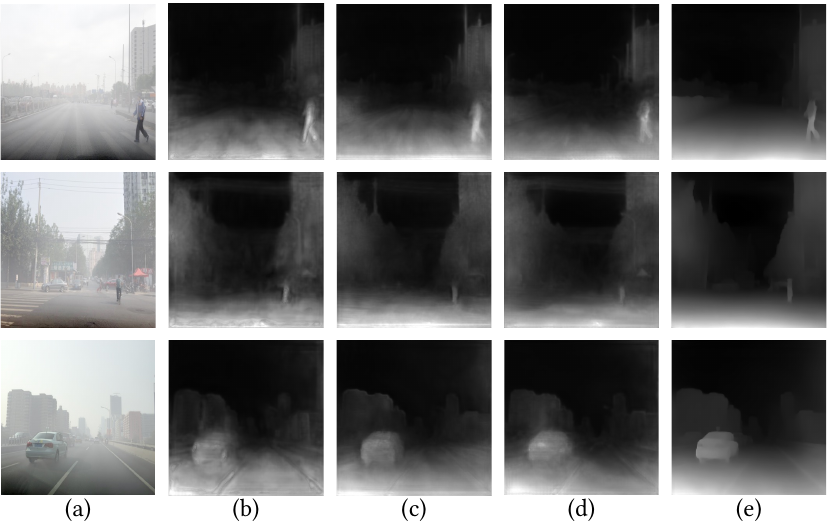}
    \caption{
    Visualization of transmission map produced by our \emph{T-Net} utilizing our framework on Haze4k dataset~\cite{haze4k}.
    Note that each \emph{T-Net} has the same architecture, but jointly trained with different \emph{D-Net} architectures in our framework.
    We denote the architecture of each branch as the combination of \emph{D-Net} + \emph{T-Net}.
    (a) Hazy image.
    (b) GCANet + GCANet.
    (c) FFA-Net + GCANet.
    (d) DehazeFormer-B + GCANet.
    (e) Ground truth transmission map.
    }
    \label{fig:trans_vis}
    \vspace{-1em}
\end{figure}

\paragraph{\textbf{Influence of Model Capacities.}}
We studied how the size of each module of our framework could affect the other by conducting experiments on Haze4K~\cite{haze4k} dataset.
We estimate the impact of the \emph{T-Net} on the performance of \emph{D-Net} by evaluating the ones we used for benchmarking Haze4K dataset.
Note that the architecture for \emph{T-Net} is fixed in this experiment.
On the other hand, to evaluate the impact of \emph{D-Net} performance on the \emph{T-Net} structure, we modified the number of filters in the hidden layer of GCANet. Specifically, we fixed the architecture of \emph{D-Net} as GCANet and employed another GCANet as \emph{T-Net} with 48 and 96 filters per layer, respectively, while the original structure has 64 filters per layer.
The estimated transmission maps and their evaluation metrics (mean squared error (MSE) and SSIM) on the Haze4K test set are presented in Figure~\ref{fig:trans_vis} and Table~\ref{tab:t-result}, respectively.
Although the same network architecture is used for \emph{T-Net}, improving \emph{D-Net} architecture from a simple one (\eg, GCANet) to advanced ones (FFA-Net, DehazeFormer-B) tends to improve the transmission accuracy.
Likewise, we observe that the performance of \emph{D-Net} improves as the capacity of \emph{T-Net} increased as reported in Table~\ref{tab:num_filters}.
In other words, \emph{T-Net} improves as \emph{D-Net} architecture improves, owing to our proposed framework that allows cooperation between two branches.
Thus, improving \emph{D-Net} better facilitates the training of \emph{T-Net} or vice versa.

We believe this is because \emph{D-Net} and \emph{T-Net} are complementary in that they are jointly trained to minimize the objective function.
Specifically, more accurate estimation of either $z$ or $\tau$ results in a lower loss value, allowing for more accurate gradient computation.
Thus, transmission estimation module can improve with dehazing module.
The results also allude to the efficacy of our joint optimization of clean haze-free image and transmission map, which are related by our proposed Bayesian framework and objective.

\begin{figure}[t!]
    \centering
    \includegraphics[width=0.5\textwidth]{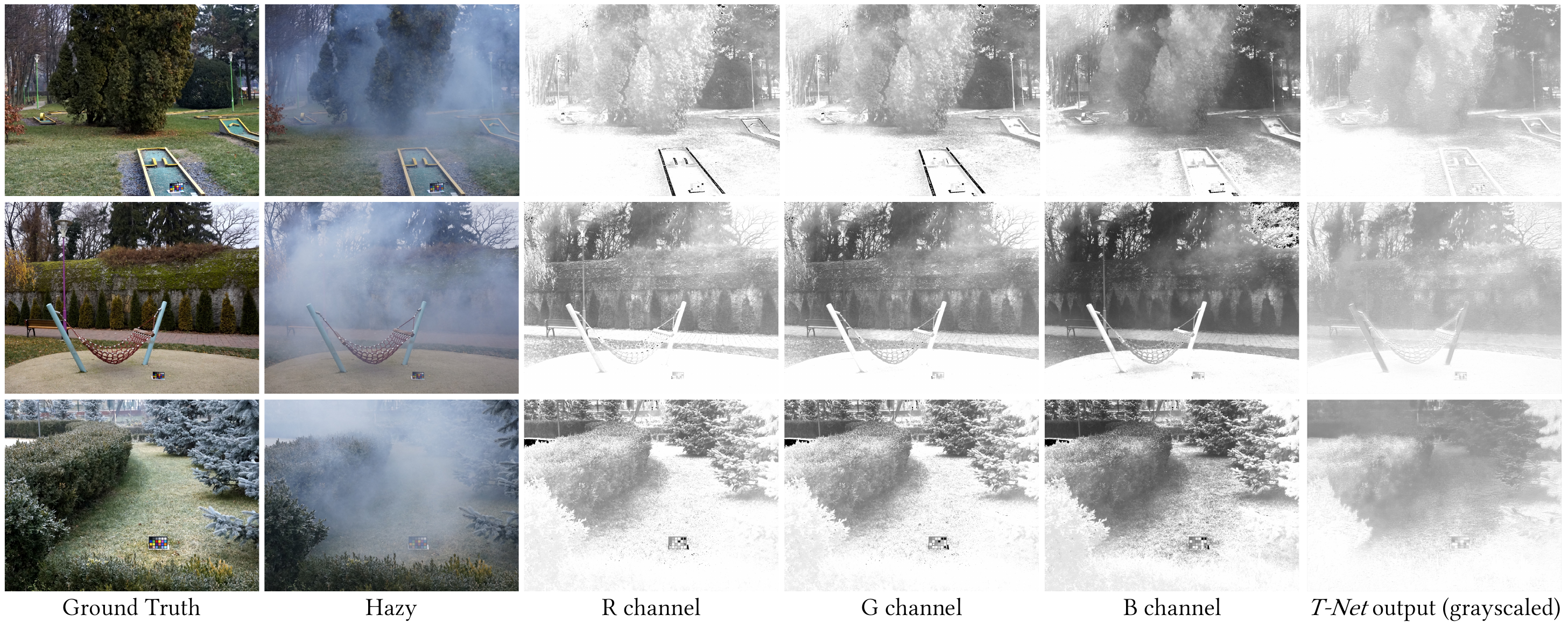}
    \caption{
    Visualization of the transmission map implemented as explained in Section~\ref{sec:tnet_explained}.
    The prior assumption is violated as the ratio differs by RGB channels.
    }
    \label{fig:nh_rgb}
    \vspace{-2em}
\end{figure}

\paragraph{\textbf{Implementation for NH-Haze Dataset.}}
Unlike synthetic datasets, the assumption of the atmosphere scattering model in Eq. (1) (\ie, $I=J\odot t + A\cdot (1-t)$) may not hold true in real-world hazy images due to several violations of its underlying assumptions.
One of the most critical violations is the significant difference in interpolation ratio among RGB channels as depicted in Figure~\ref{fig:nh_rgb}.
Note that the hazy image is represented through interpolation between the pixel value of the clean image and atmospheric light, while the transmission being the ratio between them.
Another violation is that the atmospheric light is not close to 1, leading to the interpolation ratio being outside the range between 0 and 1.
These violations make it challenging to model the haze degradation process with the atmosphere scattering model and integrate it into our framework.
However, we can mitigate the adverse effects of these assumption violations through setting $t=(I-A)/(J-A+\epsilon)$ and $A=1$.
By doing so, we can model different interpolation ratios between 0 and 1 for each RGB channel under this assumption.
As a result, this implementation enables modeling the haze degradation process on real scenarios, including depth-independent scenarios, and integrating the atmosphere scattering model into our framework.
Finally, as detailed above, \emph{D-Net} can leverage the outputs of \emph{T-Net} to enhance its dehazing performance.

\begin{table}
  \centering
  \begin{tabular}{c c c c}
    \toprule
    $z$ distribution & $\tau$ distribution & PSNR$\uparrow$ & SSIM$\uparrow$ \\
    \midrule
    Gaussian & Gaussian & 30.96 & 0.9748 \\
    Gaussian & Lognormal & 31.22 & 0.9772 \\
    Laplace & Gaussian & 31.16 & 0.9771 \\
    Laplace & Lognormal & \textbf{31.25} & \textbf{0.9793} \\
    \midrule
    \multicolumn{2}{c}{Baseline (GCANet~\cite{chen2018gated})} & 30.77 & 0.9777 \\
    \bottomrule
  \end{tabular}
  \caption{
  The PSNR(dB), SSIM results of GCANet~\cite{chen2018gated} + Ours on SOTS-Indoor dataset~\cite{li2018benchmarking} according to changing the prior models.
  }
  \label{tab:distribution}
  \vspace{-1em}
\end{table}


\begin{table}
    \centering
    \begin{tabular}{c c c c}
    \toprule
    Model & GCANet~\cite{chen2018gated} & FFA-Net~\cite{qin2020ffa} & DF-B~\cite{song2022vision} \\
    \midrule
    MSE$\downarrow$ & 0.028 & \textbf{0.019} & 0.020 \\
    SSIM$\uparrow$ & 0.711 & \textbf{0.770} & 0.747 \\
    \bottomrule
    \end{tabular}
    \caption{
    The MSE, SSIM results of inferenced transmission map on Haze4K test set~\cite{haze4k}.
    Both metrics improve as the performance of \emph{D-Net} on Haze4K benchmark increase.
    }
    \label{tab:t-result}
    \vspace{-1em}
\end{table}


\begin{table}
    \centering
    \begin{tabular}{c c c c}
    \toprule
    \# of filters & 48 & 64 & 96 \\
    \midrule
    PSNR$\uparrow$ & 24.72 & 25.28 & \textbf{25.74} \\
    SSIM$\uparrow$ & 0.9349 & 0.9399 & \textbf{0.9447}\\
    \bottomrule
    \end{tabular}
    \caption{
    The performance of \emph{D-Net} in terms of PSNR and SSIM on Haze4K test set.
    Both metrics improved as the capacity of \emph{T-Net} increased.
    }
    \vspace{-2em}
    \label{tab:num_filters}
\end{table}


\section{Conclusion}
\label{sec:conclusion}
This work is founded on the motivation that there are inherent uncertainties that make the single image dehazing problem challenging.
To alleviate this problem, we propose to formulate a variational Bayesian framework for single image dehazing.
Incorporating the atmospheric scattering model, we handle uncertainties involved in estimating transmission and haze-free images.
In particular, we take transmission and haze-free images as latent variables and use neural networks to parameterize the approximate posterior distribution of these joint latent variables.
Our framework provides consistent performance improvement across various models and numerous datasets.

\section*{Acknowledgement}
This work was supported by Institute of Information \& communications Technology Planning \& Evaluation (IITP) grant funded by the Korea government(MSIT) (No.2022-0-00156, Fundamental research on continual meta-learning for quality enhancement of casual videos and their 3D metaverse transformation) and
Institute of Information \& communications Technology Planning \& Evaluation (IITP) grant funded by the Korea government(MSIT) (No.2020-0-01373, Artificial Intelligence Graduate School Program(Hanyang University))
\balance

\bibliographystyle{ACM-Reference-Format}
\bibliography{main}

\end{document}